# Epistemological Relevance and Statistical Knowledge[s]

Henry E. Kyburg, Jr. University of Rochester
Rochester, N. Y. 14620

## 1. Background.

For many years, at least since McCarthy and Hayes (1969), writers have lamented, and attempted to compensate for, the alleged fact that we often do not have adequate statistical knowledge for governing the uncertainty of belief, for making uncertain inferences, and the like. It is hardly ever spelled out what "adequate statistical knowledge" would be, if we had it, and how adequate statistical knowledge could be used to control and regulate epistemic uncertainty.

One response to the lack of adequate statistics has been to search for non-statistical measures of uncertainty. The minimal variant has been to propose "subjective probability" as a concept to which we can turn when we lack statistics.

This proposal comes in widely differing flavors, corresponding to the dreadful ambiguity of "subjective". Sometimes this means merely "indexed by a subject". In this sense there is no conflict with statistical representations: the "subjectivity" involved just represents the fact that statistical knowledge is related to (had by) a knower. (This appears to be the sense of "subjective" employed by Cheeseman (1987).)

At the other extreme, "subjective" may mean arbitrary, subject to no objective control or constraint. Those who think we must turn in this direction are influenced by the feeling that in many cases there may be nothing better to turn to. The philosopher F. P. Ramsey, who did much to make the subjective approach to uncertainty respectable, wrote: "...a man's expectation of drawing a white or a black ball from an urn ... may within the limits of consistency be any he likes..." (1931, p. 189).

Other proposals concern non-probabilistic measures of uncertainty: the certainty factors of MYCIN (1984), the belief functions of Shafer (1987), the fuzzy membership relation of Zadeh (1986).

Our purpose here is not to evaluate these alternative treatments of uncertainty, but rather to explore the question of how far you can go on the basis of statistical knowledge that you do have, and what considerations must be taken account of in this attempt. Relatively few people have explored the question of how far you can go using statistical knowledge.[1]

A second question, in fact the one that McCarthy and Hayes had in mind, is the question of using statistical knowledge to provide an underpinning for uncertain inference -- that is, inference that is based on incomplete knowledge. A basic presupposition of the non-monotonic and
default industries seems to be that you cannot very often base such
inferences on statistical knowledge. In part this presupposition is based on
the feeling that "typicality" and "frequency" mean different things. Be that
as it may, in formalizing non-monotonic logics many people seem to be led
to considerations that (not surprisingly!) mirror considerations appropriate
to the application of statistical knowledge.

Thus Etherington (1987) introduces the concept of preference
among models; Konolige (1987) defines a notion of minimal extension;
Touretzky (1986) gives a metric for inferential distance.

These ideas will be reflected in the principles governing the
relvance of statistical knowledge to be discussed below. Our analysis of the
ground-rules for the use of statistical knowledge will throw light on the
"cancellation principles" of non-monotonic logic as well.

## 2. Assumptions.

The assumptions we make here are four.

(1) We suppose that the knowledge base may have objective statistical
knowledge in it. This statistical knowledge may be construed in a number
of ways -- for example as statements concerning chances or statements
concerning frequencies in an arbitrarily long run, or statements concerning
frequencies summed over possible worlds. We do suppose that these
statements are general: that is that they do not represent the fact that we
have recorded a frequency in a specific sample. We may have done so, and
gone on to infer a general statistical statement. But we also may have
gotten our statistical knowledge from a handbook, or a dependable
colleague. In any event, the statistical knowledge in our knowledge base is
taken to be general scientific knowledge relating properties; we will write
"$\%(A,R) = p$" to represent the fact that the long-run frequency $A$'s among
$R$'s is $p$.[1] We will weaken this assumption later to take account of
approximate statistical knowledge. The long run frequency of $A$'s may be
about p, or at least q.

(2) We assume that our body of knowledge -- our background
knowledge -- determines equivalence classes of the statements whose
uncertainty concerns us. $S$ and $T$ are in the same equivalence class if we
know that they have the same truth values: that is, if and only if the truth
functional biconditional $S \square T$ is part of our knowledge. Thus if we know
that the next toss of this coin is the next toss of a 1979 U.S. quarter and
that it lands heads if and only if it fails to land tails, and that I will choose
to have chocolate ice cream if and only if it lands tails, then

"The next toss of this coin lands heads"

238

"The next toss of a 1979 U.S. quarter lands heads"
"The next toss of this coin does not land tails"
"I will not choose chocolate ice cream"
all fall in the same equivalence class and have the same probability.

The <u>question</u> to which we take statisical knowledge to be relevant is the whole equivalence class of statements. It is this assumption that ensures that we always have statistical knowledge that bears (relevantly) on any given statement, however unique[2] the subject of that statement. It is also this assumption that requires us to think about the principles of statistical relevance: if "$S \square T$" is in our knowlege base, then the same statistical knowledge that is relevant to $S$ must also be relevant to $T$.

(3) We assume, as usual, that our knowledge base can be expressed in a first order extensional language.[3] We may still take an individual to be arbitrarily complex: for example it might be a trial of a complicated compound experiment -- say an ordered triple consisting of a room in a house, an urn in the room, and a ball in the urn.

(4) Finally, in order for statistics to be of interest, we must suppose that we may know some things about an individual without knowing everything about it. Thus we might know of "the next trial" that it is a trial consisting of selecting one of a number of urns at random, and then selecting a coin at random from the urn, and then tossing the coin 10 times. And then we might be interested in whether the tenth toss landed heads on that trial, or we might know the number of heads among the tosses, and we might be interested in whether the urn was urn number 4, or we might be interested in knowing something about the frequency of two headed coins in the urn from which we got our sample.

## 3. Interference I.

We will be concerned with the way in which some items of statistical knowledge can interfere with the epistemic relevance of other items.

If all we know of Tweety is that she is a bird, it is reasonable to believe that she can fly. If we also know that she is a penguin, then it is reasonable to believe that she cannot fly, since our knowledge about the chances of a penguin flying <u>interferes</u> with our knowledge about the chances of a bird flying. If (as we may in our biological ignorance suppose) there is a rare kind of penguin that <u>can</u> fly, and if we know that Tweety is one of them, then this new knowledge interferes with our general knowledge about penguins, and again we may suppose that Tweety <u>can</u> fly.

This relation has been noted by Etherington, Poole, Konolige, and others. It corresponds to what Reichenbach (1949) had in mind when he said that we should base our "posits" (degrees of belief) on the "narrowest" reference class concerning which we have adequate statistics.

A principle embodying this natural constraint must be stated with somewhat more generality than is at first obvious, however.

Suppose (to move to an artificial example) we know of ball # 18 that it is a ball in certain room, and that we know that fifty percent of the balls in that room are black. Suppose we know also that that particular ball is also one in an urn, urn A, in which 80% are black. The second piece of statistical knowledge is clearly epistemically relevant and the first is not. This intuition is based on the fact that the set of balls in the urn in the room is a subset of the set of balls in the room.

But maybe we want to know whether *the next ball to be chosen from the room* is black. Under the usual conditions of such hypothetical experiments, we are entitled to infer that 50% of the choosings of balls are choosings of black balls. So we may consider the frequency with which such choosings are choosings of black balls.

This should be irrelevant since we know that # 18 came from an urn in which 80% of the balls are black. But the set of balls in that urn isn't a set of choosings at all and so not a subset of the set of objects we are now considering. The subset principle is of no direct help to us. Nor should it be, for it is certainly intelligible to suppose that while the frequency of black choosings corresponds to the proportion of black balls in general, it fails to do so in the set of choosings from urn A. If this is not the case -- if we have reason to believe that 80% of the choosings from urn A yield a black ball -- then either structure should be acceptable, despite the fact that one concerns choosings (events) and the other balls(objects).

We could stipulate that all the sentences in question have some specific canonical logical form; but we shall see shortly that that is not such a good idea. What we can do instead is to formulate our principle a bit more broadly:

The Subset Principle: Suppose that "$a$ is a $B$" is in our knowledge base, and that "$\%(C,B) = p$" is in our knowledge base. Suppose that we know that $a'$ is a $C'$ if and only if $a$ is a $C$. The statistical knowledge that $a'$ is a $B'$, and that $\%(C',B') = p'$, where $p \square p'$, is *epistemically irrelevant* if we know of a subset of $B'$, $B''$, such that we know both $a'$ is a $B''$ and $\%(C',B'') = p$.

The subset principle is one that has been frequently identified in the context of non-monotonic logic.

4. **Mutual Interference.**

The subset principle does not always help, as the Nixon Diamond indicates. (Again we see a relation between the application of statistics and non-monotonic logic.) Suppose person #18 is known to be a quaker, and that we know that 90% of quakers are pacifists; and also that person #18 is a republican, and that 20% of republicans are pacifists. It seems clear that neither the 20% nor the 90% are the relevant frequencies. If we knew the proportion of pacifists among republican quakers, we could usefully apply the subset principle, but we have no reason to suppose that we know this proportion (which, of course, may have any value from 0 to 1). If we allow vague statistical knowledge, we will have knowledge about the intersection, but it might be only that the proportion of pacifists is somewhere between 0% and 100%.

It is not entirely unreasonable that conflicting information can leave us in a state of ignorance. But sometimes we feel that we can do bettter. There are a number of ways in which we can do better, though no completely general procedure seems credible.

## 5. Interference II.

Here is an example that calls for a second principle: As before, suppose we have a roomful of urns, and that #18 designates a ball in the room. Suppose we know that there are 100 balls in the room, and that 50 are black. But suppose we also know that there are 10 urns, that 9 of them containing four black balls and one white ball, and that the tenth contains the remainder of the balls. The relative frequency of black balls in the first nine urns is .80, and the relative frequency of black balls in the tenth urn is $14/55 = .25...$

Let us consider what statistics are relevant to the statement, "#18 is black." If we know of #18 only that it is a ball in the room, it is only the statistics about the frequency of black balls in the room that are relevant. If we know also something about how #18 came to be the designated ball, the other statistics may also be relevant. For example, we might know that #18 is the ball resulting from first choosing an urn at random, and then choosing a ball at random from the chosen urn. If that is the case, the relevant statistics are those governing the proportion of pairs consisting of an urn, and a ball drawn from that urn, such that the second member of the pair is black. We can easily calculate the proportion of pairs having this property to be $.9 * .8 + .1 * .25... = .745...$

But why, under these circumstances, should we regard the general statistics concerning balls in the room to be epistemically irrelevant? The interfering set isn't a subset of its competitor. (Note that .745... cannot, mathematically, be the relative frequency in any subset of the set of balls in the room!)

But we can find a relationship: there is a possible reference class that matches the competitor, of which the correct reference set is a subset -- namely, the cross product of the set of urns and the set of balls. This construction is particularly important in the context of (so-called) Bayesian inference; the model we just looked at corresponds to a non-sampling case in which we have a prior probability of .9 combined with a conditional probability of .8, and a prior probability of .1 combined with a conditional probability of .25... We therefore call the rule the Bayesian Principle:

<u>The Bayesian Principle:</u> Suppose that "$\langle a,b \rangle$ is a $B$" is in our knowledge base, and that "$\%(C,B) = p$" is in our knowledge base. The statistical knowledge that $a'$ is a $C''$ if and only if $a$ is a $C$, that $a'$ is a $B'$, and that $\%(C',B') = p' \;\square\; p$ is *epistemically irrelevant* if we know of a cross product of $B'$ with $B''$ and a corresponding subset $C''$ and $a''$ such that
        (1) $\langle a',a'' \rangle$ is known to be in $B' \diamond B''$,
        (2) $\langle a',a'' \rangle$ is in $C''$ if and only if $a$ is in $C$,
        (3) $\%(C'', B' \diamond B'') = p'$,
and for some $B^*$ known to be a subset of $B' \diamond B''$,
        (4) $\%(B^*, C) = p$.

The Bayesian principle is followed in constructing representations of uncertainty in which uncertainties are modified by new evidence, but I have not noticed it in discussions of non-monotonic inference. It should be, of course.

Almost all (species of) mammals give birth to their young live. Given an arbitrary individual mammal, we do not have reason to think it gives birth to its young live, since half of mammals don't give birth at all. (The analogous point with respect to birds was pointed out by Nutter (1987).) Given an arbitrary individual <u>female</u> mammal, we have reason to think it will give birth to its young live, since almost all species of mammals are such that when their females reproduce, they do it that way. (Almost all the reproductive balls in almost all the urns are white, though it is not the case that almost all the balls in an urn are reproductive.) We accommodate the ovoviviparous platypus by noting that its <u>species</u> (its urn) is unusual.

## 6. Interference III.

The final principle of relevance we need for dealing with statistical knowledge is in a sense the dual of our first principle, the subset principle. Suppose that we are sampling from a population $C$ with a view to making an inference about the proportion of $B$'s there are in $C$. It is a general set theoretical fact that almost all subsets of a given set reflect within narrow limits the composition of the parent set. It is consistent with this observation that we should have observed more $C$'s than we mentioned explicitly. This should render the previous numbers irrelevant. The larger sample is the one that is epistemically relevant. A principle that captures this intuition is:

<u>The Supersample Principle:</u> Suppose that we know that $a_n$ is a member of $P^n$ and that we are interested in the chance that $a_n$ is $R_\square$. The statistical knowledge that $a_m$ is known to be a member of $P^m$, that $a_m$ is a $R'$ if and only if $a_n$ is $R_\square$, and that $\%(R_\square, P^n) = p \square p' = \%(R', P^m)$ is *epistemically irrelevant* if we also know that $a_m$ is a subset of $a_n$.

## 7. Discussion.

It is my belief that these three principles are all the principles we need to determine the epistemic relevance of statistical knowledge in the case in which we either have exact knowledge or none at all. It may be that other principles are needed, but I have seen no examples that intuitively require additional principles.

Should they be called "principles?" It seems to me that they should, and that they are all roughly on a par, even though the subset principles is derivable from the Bayesian principle. They may be construed collectively as an articulation
of our intuitive ideas about "total evidence."

The first principle directs us to use the most specific evidence at hand. The second directs us to take account of general background knowledge. The third says not to ignore available data. Stated thus they seem sensible enough.

## 8. Inexact Knowledge.

By providing a new characterization of "difference" among statistical statements, we can easily generalize the above considerations. Let "$\%(A,B) \in [p,q]$" *differ* from "$\%(C,D) \in [r,s]$" just in case neither of $[p,q]$ nor $[r,s]$ is included in the other. Let us say that the former is *stronger* than the latter if $[p,q] \subset [r,s]$. Then we shall say that one item of statistical knowledge is irrelevant to another if

(a) it differs, but is rendered irrelevant by one of the three principles expounded above, or

(b) it is weaker than the other and no third item of statistical knowledge renders it irrelevant.

## 8. Conclusions:

(1) If we accept the equivalence condition -- that statements connnected in our knowledge base by a biconditional should have the same probability -- then many more statements than might at first have been thought can have probabilities based on statistical background knowledge.

(2) There are three intuitive ways in which conflict between two potential reference classes can be resolved to the benefit of one of them.

(3) These three resolutions reflect the three principles: the Subset Principle, the Bayesian Principle, and the Superset Principle.

(4) The results of this analysis can be used to implement probabilistic non-monotonic acceptance as well as to determine rationally allowable distributions of uncertainty.

* Research underlying this paper has been partially supported by the Signals Warfare Center of the U. S. Army.

1. One writer who has taken this question seriously is Bacchus (1988).
2. "however unique": This way of talking is offensive to anyone with grammatical sensibilities, but so it goes.
3. Of course this requires enough arithmetic, in first order form, to accommodate the statistics!